\documentclass[review]{elsarticle}
\usepackage{graphicx}
\usepackage{subcaption}
\usepackage{algorithm}
\usepackage{algorithmic}
\usepackage{amsmath}
\usepackage{enumitem}
\usepackage{graphicx}
\usepackage{amsmath}
\usepackage{amssymb}
\usepackage{booktabs}
\usepackage{subcaption}
\usepackage{color}
\usepackage{bm}
\usepackage{multirow}
\usepackage[fleqn]{nccmath}
\usepackage[%
  colorlinks = true,
  citecolor  = RoyalBlue,
  linkcolor  = RoyalBlue,
  urlcolor   = RoyalBlue,
  unicode,
  pagebackref
  ]{hyperref}
\usepackage[capitalize,noabbrev]{cleveref}
\usepackage{selectp}

\usepackage[T1]{fontenc}    % use 8-bit T1 fonts
\usepackage{url}            % simple URL typesetting
\usepackage{booktabs}       % professional-quality tables
\usepackage{amsfonts}       % blackboard math symbols
\usepackage{nicefrac}       % compact symbols for 1/2, etc.
\usepackage[usenames,dvipsnames]{xcolor}
\usepackage{microtype}      % microtypography

\journal{Journal of \LaTeX\ Templates}
\usepackage{multirow}

\newcommand{\x}{\mathbf{x}}
\newcommand{\y}{\mathbf{y}}

\newcommand{\D}{\mathcal{D}}
\newcommand{\F}{\mathcal{F}}
\newcommand{\A}{\mathcal{A}}

\begin{document}

\begin{frontmatter}
\def \ali#1{\textcolor{blue}{#1}}

% misc
\newcommand{\cmark}{\ding{51}}%
\newcommand{\xmark}{\ding{55}}%
\newcommand{\ouralg}{SG-MRL}
\def\x{\boldsymbol{x}}
\def\r{\boldsymbol{r}}
\def\w{\boldsymbol{w}}

\DeclareMathOperator*{\argmin}{arg\,min}
\DeclareMathOperator*{\argmax}{arg\,max}
\DeclareMathOperator*{\sign}{sign}

\newcommand{\hypothesisincorrect}[2]{%
\begin{tcolorbox}[colback=gray!10!white,leftrule=2.5mm,size=title]
\textbf{#1}: #2
\end{tcolorbox}
\vspace{-0.1cm}%
}
\newcommand{\hypothesiscorrect}[2]{%
\begin{tcolorbox}[colback=gray!10!white,leftrule=2.5mm,size=title]
\textbf{#1}: #2
\end{tcolorbox}
\vspace{-0.1cm}%
}
\newcommand{\finding}[2]{%
\begin{tcolorbox}[colback=yellow!10!white,leftrule=2.5mm,size=title]
\textbf{#1}: #2
\end{tcolorbox}
\vspace{-0.1cm}%
}
\newcommand{\rjenatton}[1]{{\color{cyan} \textbf{rjenatton}: #1}}
\newcommand{\rjenattonnoname}[1]{{\color{orange}#1}}
\newcommand{\efi}[1]{{\color{violet} \textbf{efi}: #1}}
\newcommand{\guille}[1]{{\color{olive} \textbf{guille}: #1}}

\newcommand{\orig}[1]{{\color[HTML]{AF1E1E} #1}}
\newcommand{\ind}[1]{{\color[HTML]{00D59D} #1}}
\newcommand{\labs}[1]{{\color[HTML]{3492C7} #1}}
\newcommand{\nearopt}[1]{{\color[HTML]{BF9907} #1}}
\newcommand{\gray}[1]{{\color{gray} #1}}
\newcommand{\nopi}[1]{#1}

\definecolor{codegreen}{rgb}{0,0.6,0}
\definecolor{codegray}{rgb}{0.5,0.5,0.5}
\definecolor{codepurple}{rgb}{0.58,0,0.82}
\definecolor{backcolour}{rgb}{0.95,0.95,0.92}  

\title{ Fast Classification with Sequential Feature Selection in Test Phase }
% \tnotetext[mytitlenote]{Fully documented templates are available in the elsarticle package on \href{http://www.ctan.org/tex-archive/macros/latex/contrib/elsarticle}{CTAN}.}

%% Group authors per affiliation:
%%% Group authors per affiliation:
%\author{Ali Mirzaei}
%\address{Amirkabir University of Technology}
%\fntext[myfootnote]{Since 1880.}

%% or include affiliations in footnotes:
\author{Ali Mirzaei}
\ead{ali\_mirzaei@aut.ac.ir}
\author[]{Vahid Pourahmadi\corref{sdd}}
\ead{v.pourahmadi@aut.ac.ir}
\cortext[sdd]{Corresponding author}
\author{Hamid Sheikhzadeh}
\ead{hsheikh@aut.ac.ir}
\author{Alireza Abdollahpourrostam}
\ead{alirezaap\_r@aut.ac.ir}

%\ead{support@elsevier.com}

\address{Amirkabir University of Technology, Tehran, Iran}

%\address[ali]{Amirkabir University of Technology,Hafez Avenue, Tehran, Iran}

% %% or include affiliations in footnotes:
% \author[Ali Mirzaei, Vahid Pourahmadi]{Amirkabir University of Technology, Tehran, Iran}
% \ead[url]{www.elsevier.com}

% \author[Vahid Pourahmadi]{Global Customer Service\corref{mycorrespondingauthor}}
% \cortext[mycorrespondingauthor]{Corresponding author}
% \ead{support@elsevier.com}

\begin{abstract}
This paper introduces a novel approach to active feature acquisition for classification, which is the task of sequentially selecting the most informative subset of features to achieve optimal prediction performance during testing while minimizing cost. The proposed approach involves a new lazy model that is significantly faster and more efficient compared to existing methods, while still producing comparable accuracy results.

During the test phase, the proposed approach utilizes Fisher scores for feature ranking to identify the most important feature at each step. In the next step the training dataset is filtered based on the observed value of the selected feature and then we continue this process to reach to acceptable accuracy or limit of the budget for feature acquisition. The performance of the proposed approach was evaluated on synthetic and real datasets, including our new synthetic dataset, CUBE dataset and also real dataset Forest. The experimental results demonstrate that our approach achieves competitive accuracy results compared to existing methods, while significantly outperforming them in terms of speed. The source code of the algorithm is released at github with this link: \url{https://github.com/alimirzaei/FCwSFS}.
\end{abstract}


\end{frontmatter}

% \linenumbers

\section{Introduction}

Traditional classification approaches assume all features of data are freely available and can be used in the process of training and testing. However, in some real problems, acquiring of all features are costly and this cost could be time, financial cost, risk of acquisition, energy consumption. In such problems, the goal is the highest accuracy while spending the least cost (using least features). For example, when a patient goes to a doctor, the doctor first measures the symptoms and features that are easy to measure and have a low measurement cost. Then, based on the observations of those values, as well as the knowledge and experience that he has, he guesses about the person's illness.  According to his guess, he determines and prescribes the next features and tests that the patient should measure and this procedure can be repeated to the doctor can diagnose the disease correctly. Obtaining attributes in this way and actively is very useful and important for applications that acquisition of features are difficult and costly in terms of time, money and health risk. To this end, an AI model should be able to suggest the features to be acquired subsequently for diagnosis of new patients. Such model also can be employed in other application which the acquisition or transition of data are costly.

This paper proposes a fast and efficient statistical approach that addresses the limitations associated with existing methods. Particularly, the proposed approach is highly effective for datasets with limited samples or those that undergo frequent changes in training data. This is especially relevant as existing approaches often require retraining when the size of the training sample changes, which can be computationally intensive and impractical in practice. 

The paper presents several significant contributions:

\begin{itemize}
    \item Introduction of a novel and general approach for active feature acquisition in classification. This approach allows for the utilization of any feature ranking as the central component of the overall methodology. The choice of the feature ranking approach can be tailored to the specific dataset and application requirements.

    \item Proposal of the Fisher scoring approach as a suitable method for real-time and general use cases. The use of Fisher scoring enhances the effectiveness and efficiency of feature acquisition in various scenarios.

    \item Introduction of a new synthetic dataset designed to evaluate the proposed methods from different perspectives. This dataset is specifically constructed to incorporate varying levels of feature importance during the classification process, as well as considering the sequential order in which features are measured.
\end{itemize}

\section{Related Works}

Current existing approaches for costly feature classification are classified into three general classes: \\
\textbf{Decision Tree based Approaches}: These methods try to solve this problem using the decision trees. In the following we listed some of these approaches:

Xu et al. (2013) presented an algorithm termed as the Cost-Sensitive Tree of Classifiers (CSTC) \cite{xu2013cost}, which is designed to construct a tree of classifiers that reduces the average test time complexity of machine learning algorithms, while simultaneously enhancing their accuracy. The fundamental concept of the CSTC algorithm proposed in the paper is a computationally complex task, as it is classified as NP-hard. To mitigate this complexity, the authors proposed a mix-norm relaxation approach to approximate the cost of measuring features. However, it is important to note that this approach requires parameter fine-tuning and the training process can be time-consuming.

Kusner et al. (2014) proposed a distinct relaxation approach using approximate submodularity, referred to as the Approximately Submodular Tree of Classifiers (ASTC) for the Cost-Sensitive Tree of Classifiers (CSTC) algorithm \cite{kusner2014feature}. The ASTC approach is comparatively more straightforward to implement, provides equivalent results to CSTC, and does not require the fine-tuning of optimization hyperparameters. Furthermore, the training process of ASTC is up to two orders of magnitude faster than CSTC.

Nan et al. (2015) proposes a novel random forest algorithm to minimize prediction error for a user-specified average feature acquisition budget, where each feature can only be acquired at an additional cost during prediction-time \cite{nan2015feature}. The algorithm grows trees with low acquisition cost and high strength based on greedy minimax cost-weighted-impurity splits. 
    
Nan et al. (2016) proposes a method to prune random forests (RF) to optimize feature cost and accuracy for resource-constrained prediction \cite{nan2016pruning}. The authors pose pruning RFs as a 0-1 integer program with linear constraints that encourages feature re-use and establish total unimodularity of the constraint set.

Nan and Saligrama (2017) train a gating and prediction model to limit the use of a high-cost model for hard inputs, and gate easy inputs to a low-cost model \cite{nan2017adaptive}. Their method adaptively approximates the high-cost model in regions where low-cost models suffice. They use empirical loss minimization with cost constraints to jointly train gating and prediction models

    \textbf{Generative Imputation Approaches}: These models uses the generative approaches to learn distribution of unknown features based the known ones. EDDI (Efficient Dynamic Discovery of high-value Inference) \cite{ma2018eddi} uses a partial variational autoencoder to predict the distribution of features and then it selects the most informative features based on obtained distribution (the features which has higher entropy will be more informative for acquisition). Ice-Breaker \cite{gong2019icebreaker} is another approach the same  bas EDDI which uses a Bayesian Deep Latent Gaussian Model to learn distribution. Since these approaches learn distribution they also could be used for imputation task (prediction other features based on measured ones) but they have high computational complexity ( in case of high number of features). \\
% \textbf{Information theoretical based Feature Selection:}
% In contrast to similarity-based feature selection algorithms, which are incapable of handling feature redundancy, information theoretically-based feature selection algorithms take both feature relevance and feature redundancy into account. The majority of these methods rely on label data to evaluate the relevance and redundancy of features, so these features are typically supervised. Minimum Redundancy Maximum Relevance~(MRMR)~\cite{peng2005feature} is an example of a well-known member of this family. The aforementioned approach chooses a feature set that is optimal by utilising the criterion of maximal statistical dependency, which is determined by mutual information. The majority of extant feature selection methods based on information theory are restricted to supervised scenarios, requiring the availability of labels. \\
\textbf{Reinforcement Learning Approaches}: These approaches model the costly feature classification as a reinforcement learning problem. In this modeling reward is defined as the accuracy of classification and punishment is defined as cost of feature acquisition. \cite{dulac2011datum} models the problem as a Markov Decision Process (MDP)  and  solve  it  with linearly approximated Q-learning. \cite{cwcf} proposes a deep reinforcement learning framework for classification tasks with costly features. The authors build upon the linearly approximated Q-learning algorithm by introducing a deep Q-network (DQN) to model the problem. Specifically, the DQN is trained to either select the next feature that should be measured or classify the instance into one of the pre-defined classes. In the former case, the DQN is penalized for the cost of the feature acquisition, while in the latter case, it is penalized for the incorrect classification. The balance between the two types of penalties reflects the level of stringency imposed on the feature acquisition process, which is regulated by a given budget constraint. \cite{chen2021costly} also provides a Deep Reinforcement Learning approach based on Monte Carlo TreeSearch method  for  cost  features  classification. When the number of features increases the the state dimension will exponentially increase and in the RL problems, high dimensional state leads long training time and the algorithm may not convert to the optimum policy. 

One significant challenge associated with the approaches discussed is the extensive training time required by the models. Specifically, as the number of features increases, the dimensionality of the actions also increases, leading to a significant increase in the size of the action space. This, in turn, prolongs the training process and requires considerable computational resources. Furthermore, designing an appropriate neural network structure that can effectively handle a given dataset can be challenging. Moreover, it is worth noting that these approaches are not suitable for small datasets.

\section{Problem Formulation}

In this section, we formally define the problem of active feature acquisition for classification. We aim to select a subset of features that maximizes prediction performance during testing while minimizing the cost associated with acquiring these features.

Let:
\begin{itemize}
\item $\D_{\text{train}}$ denote the training dataset, consisting of $N$ instances and $M$ features. Each instance is represented as a feature vector $\x_i \in \mathbb{R}^M$ and is associated with a class label $\y_i$.
\item $\D_{\text{test}}$ represent the testing dataset, which is used to evaluate the prediction performance of the selected feature subset.
\item $C$ denotes the budget for feature acquisition, representing the maximum cost that can be incurred during the acquisition process.  In the proposed approach, it is assumed that all features have the same cost. Therefore, we can interpret $C$ as the maximum number of features to be selected.
\item $\F$ be the set of all features, where $\F = {1, 2, \ldots, M}$.
\end{itemize}

We define the following notations:
\begin{itemize}
\item $S$ denotes the current subset of features selected for classification. Initially, $S$ is an empty set.
\item $f_i$ represents the $i$-th feature in the sample, where $f_i \in \F$.
\item $\A(S)$ is a function that represents the acquisition cost of the feature subset $S$. It returns the total cost of acquiring all features in $S$. 
\item $P(S)$ denotes a performance metric that evaluates the prediction performance of the classifier using the feature subset $S$.
\item $S^\ast$ represents the optimal subset of features that achieves the highest prediction performance within the given budget $C$.
\end{itemize}

The objective is to find the feature subset $S^\ast$ such that:
\begin{equation}
    S^\ast = \text{argmax}_SP(S) \text{ subject to } \A(S) \leq C(S^\ast) 
\end{equation}
The problem involves a sequential decision-making process. In the following sections, we present our proposed approach for active feature acquisition, which employs a new lazy model that improves efficiency while maintaining comparable accuracy results. 

\section{The Proposed Method}
In this section we illustrate our proposed method for the problem of active feature acquisition for classification purpose. The proposed algorithm is designed to efficiently select informative features during test time for classification purposes. It takes as input the test data set $\D_{test}$, the training data set $\D_{train}$, and the maximum budget of feature acquisition ( when the cost of all features are the equal this would be the number of features) $C_{max}$.

For each test instance $\x_t \in \D_{test}$, the algorithm selects the most informative features using ANOVA F-Scoring \cite{fisher1992statistical} based on all training dataset and then after observing the value of that feature the algorithm filters the training dataset based on the measured value of that feature. For this purpose we calculate the Euclidean distance of measured features between the test sample and all training dataset and only keep the training instances that their distance with the test sample is lower than a threshold. In the next step it does the feature selection based on the filtered dataset and select the most informative feature that has not yet been selected yet. The loop continues until either the maximum budget ( maximum number of features ) $C_{max}$ is reached or no more samples are found in the filtered training dataset.

Once the selected features have been determined, the algorithm predicts the label of the test instance $\x_t$ by assuming that the most frequent label of the filtered training labels is the true label for the test instance. 

In summary, the proposed Active Feature Acquisition during Test Time algorithm provides an effective and efficient way to select informative features during test time, while also ensuring that the selected features are relevant to the current test instance. The algorithm achieves this by iteratively selecting the most important feature, observing its value, and filtering the training set based on the observed feature value until the maximum number of features. The most frequent label of the filtered training labels is used to predict the label of the current test instance. The algorithm is detailed in algorithm \ref{algorithm1}.

\begin{algorithm}
\caption{Active Feature Acquisition during Test Time}
\begin{algorithmic}
\label{algorithm1}
\REQUIRE {$\D_{test}, \D_{train}, C_{max}, TH$} 
\FOR{$\x_t \in \D_{test}$}
    \STATE $\D_{train}^{*} \gets \D_{train}$
    \STATE $selected\_features = []$
    \WHILE{$C(selected\_features) < C_{max}$ and $|\D_{train}^{*}| > 0$}
        \STATE Compute feature importance scores using fast filtering methods like fisher scores on $\D_{train}^{*}$ and select the most prominent feature that is not selected $f_i$
        \STATE Append $f_i$ to $selected\_features$
        \STATE compute distances between selected features values $\x_t[selected\_features]$ and $\D_{train}[:, selected\_features]$ as array $R$
        \STATE Filter training set and update the $\D_{train}^{*}$ as $\D_{train}^{*}=\D_{train}^{*}[R < TH]$
    \ENDWHILE
    \STATE Assume the most frequent label of $\D_{train}^{*}$ to $\y_{test}$ as sample label 
\ENDFOR
\STATE \textbf{return} $\y_{test}$
\end{algorithmic}
\end{algorithm}

\section{Evaluation}
\label{Experiments}
In this section we compare our method with two state-of-the-art reinforce learning based methods, CWCF~\cite{cwcf} and Set-Encoding~\cite{shim2018joint}. Reinforcement learning-based active feature acquisition approaches have shown promising results in terms of accuracy on large datasets. However, these models have two significant drawbacks: The training of these models takes lots of resource and time, and these methods need huge amount of training data. These drawbacks are making them impractical for use in real-world applications. To address this issue, we have developed a simple and fast algorithm for feature selection that can be implemented efficiently and is more practical for small datasets. By leveraging the strengths of our algorithm, we can achieve accurate feature selection results without sacrificing speed and practicality. We have conducted experiments on a synthetic well-designed dataset, CUBE, and Forest datasets.

\subsection{Proposed Synthetic dataset}

\begin{figure}
     \centering
     \begin{subfigure}[b]{0.4\textwidth}
         \centering
         \includegraphics[width=\textwidth]{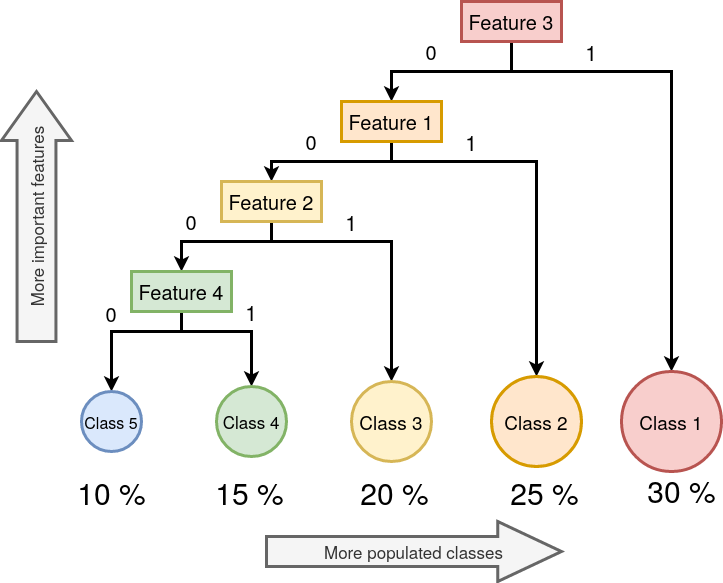}
         \caption{date generation policy}
         \label{fig:policy}
     \end{subfigure}
     \hfill
     \begin{subfigure}[b]{0.45\textwidth}
         \centering
         \includegraphics[width=\textwidth]{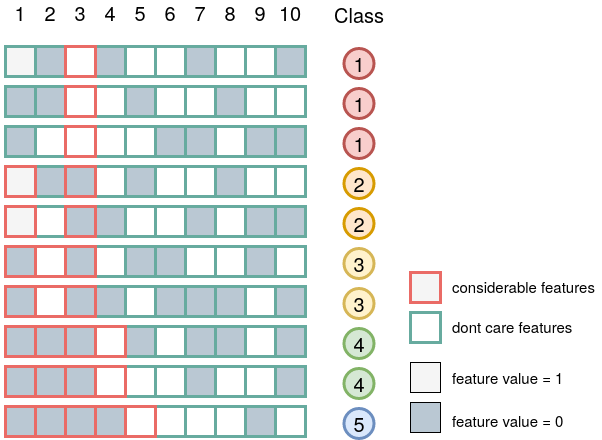}
         \caption{samples}
         \label{fig:samples}
     \end{subfigure}
        \caption{The data generation policy to the features have the importance order of $[3, 1, 2, 4, 5]$ and 10 samples of generated data}
        \label{fig:three graphs}
\end{figure}

To verify the capabilities of the proposed method, a novel dataset is designed such that the information of each feature is different and also the order of measuring features is important.
To this end, we generate a binary data with $10$ dimensions (features) that only the first five features are important and relevant to the instance class (label) and other features ($6^{th}-10^{th}$) are noisy and irrelevant. We assume the arbitrary order of $[3, 1, 2, 4, 5]$ for importance of measuring the relevant features. To realize this order, we generate data such that follow the policy shown in figure \ref{fig:policy}. According this policy for classifying an instance, we have to first measure the $3^{rd}$ feature and if it was equal to one then this instance classifies in class $1$. Otherwise, we have to measure the next feature and we have to continue the same procedure to visit all relevant features. Figure \ref{fig:samples} also shown some samples of our synthetic dataset.  

\subsubsection{Expectations from Algorithm}
Using this dataset we expect that the algorithm can do the following:

\begin{enumerate}
    \item At initial state (without measuring any feature), it must detect the most important feature which should be measured ( the 3rd feature )
    \item Based on the value of measured feature it must able to suggest the right order of features.
    \item It has to detect the irrelevant features and does not suggest these features.
\end{enumerate}

\subsubsection{Experiment Results}

\begin{figure}
    \centering
    \includegraphics[width=0.65\textwidth]{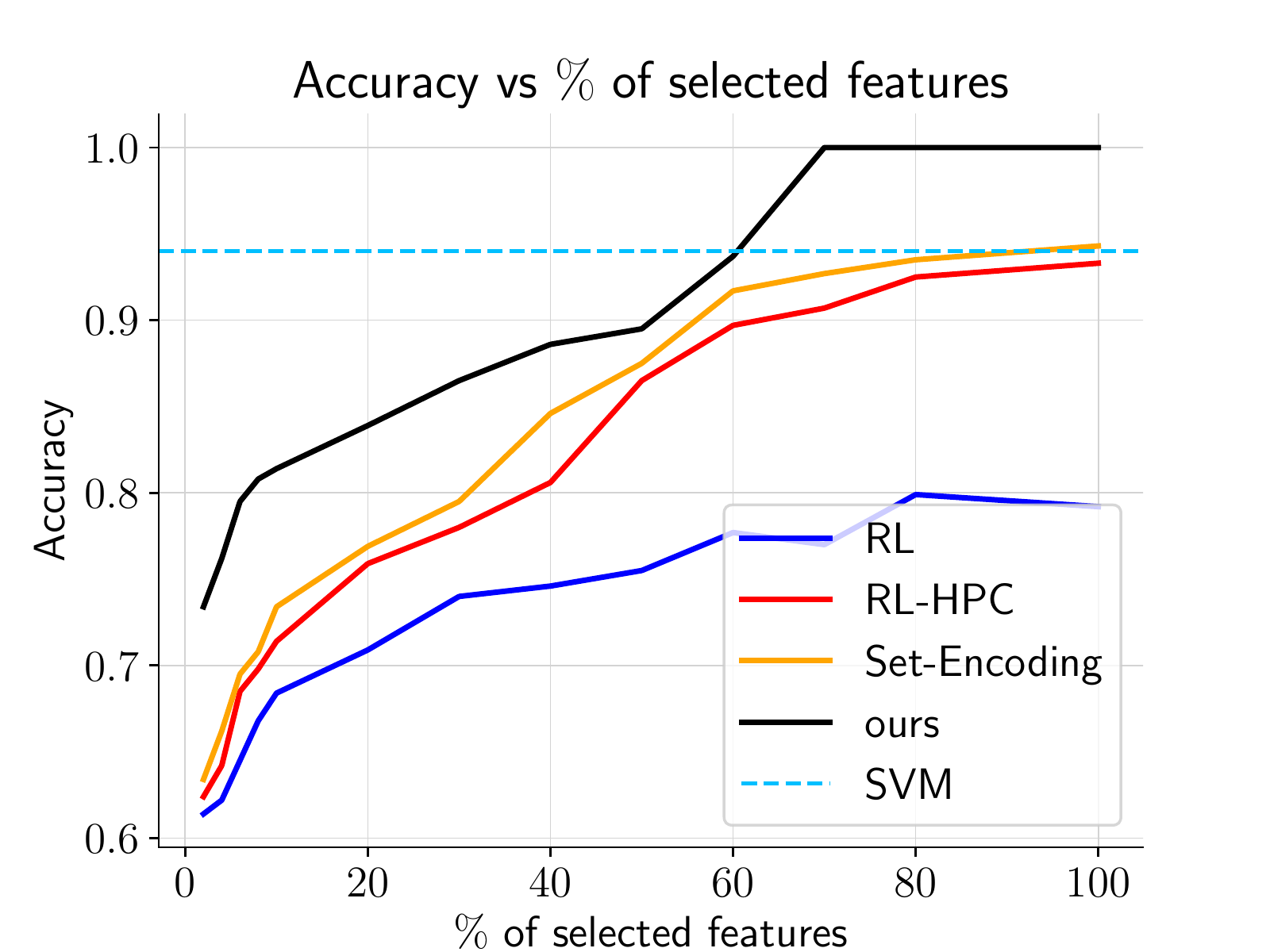}
    \caption{Our Synthetic dataset. The CWCF method is referred to as RL in this graph, while Set-encoding, another RL algorithm, is given a specific name.}
    \label{fig:ours}
\end{figure}

In this study, we conducted an experiment involving the generation of 5000 random samples for training data and an additional 100 samples for the testing dataset. To provide a more precise evaluation of the CWCF, we trained a model incorporating HPC (High-Performance Classifier). As a baseline comparison, we also included the results of an SVM (Support Vector Machine) classifier \cite{svm} utilizing all features.

The figure presented as Figure \ref{fig:ours} illustrates the accuracy of classification based on the number of measured features. It is evident that our proposed approach successfully identifies the correct sequence of features, resulting in higher accuracy compared to other approaches. Moreover, our method effectively disregards irrelevant features, as demonstrated by achieving 100% accuracy after measuring just 5 features, which constitutes more than 50% of the total features.

The observed lower performance of the CWCF, depicted as the RL method in the figure, can be attributed to several factors. Firstly, our dataset is relatively small, which can lead to overfitting of RL approaches to the training data or hinder the learning of a robust dataset representation. Consequently, this limitation negatively impacts the performance of these approaches.

\subsection{CUBE dataset} 

In the next phase, we conduct experiments using a synthetic dataset called CUBE-$\sigma$ to assess the effectiveness of our approach in identifying key features that pertain to the assigned classification task. This datset classify samples in $5$ classes and only the first $5$ features are relevant to the class label and the rest of other features are noisy features. The informative features are also added with a Guasian noise with standard deviation of $\sigma$.  To gain a deeper understanding of CUBE-$\sigma$, please refer to the work by Shim et al. \cite{shim2018joint}. 

We set the Gaussian noise parameter $\sigma$ to 0.3 for the informative features and all models are trained on $5000$ generated samples with $20$ features (only $5$ of them are informative) and $100$ samples are generated for testing purpose.

\begin{figure}
    \centering
    \includegraphics[width=0.6\textwidth]{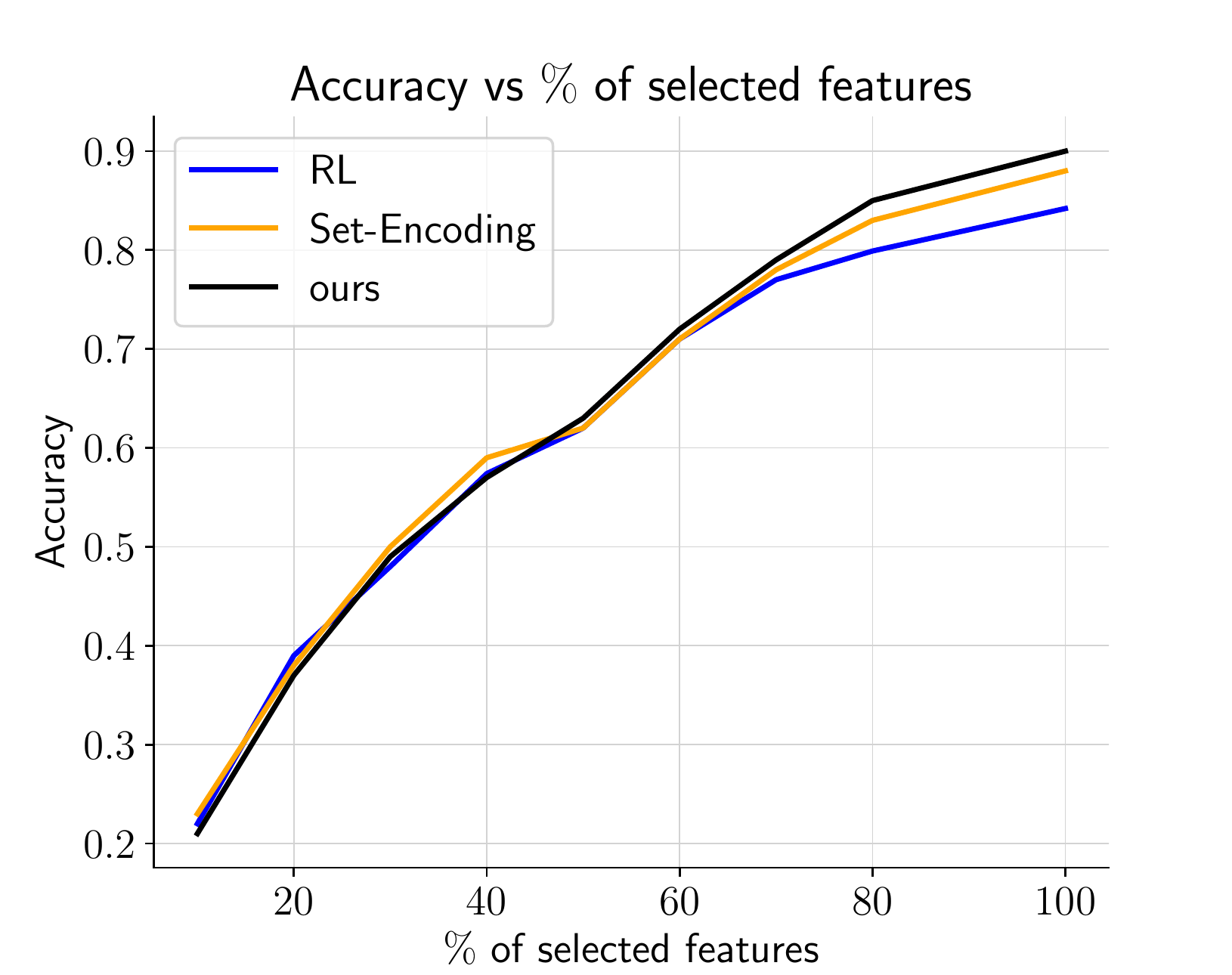}
    \caption{CUBE with $20$ features. The CWCF method is referred to as RL in this graph, while Set-encoding, another RL algorithm, is given a specific name.}
    \label{fig:cube}
\end{figure}

As shown in figure~\ref{fig:cube}, the proposed approach has a comparable results with the RL based approaches while the training time of it is zero and testing time is comparable with these methods.

An additional noteworthy observation derived from the analysis of figure~\ref{fig:cube} is that our approach exhibits a slight superiority over other reinforcement learning-based methods when the algorithm is granted access to the entirety of features and this shows the ability of the algorithm in ignoring the irrelevant features.

\subsection{Forest Dataset}

\begin{figure}
    \centering
    \includegraphics[width=0.65\textwidth]{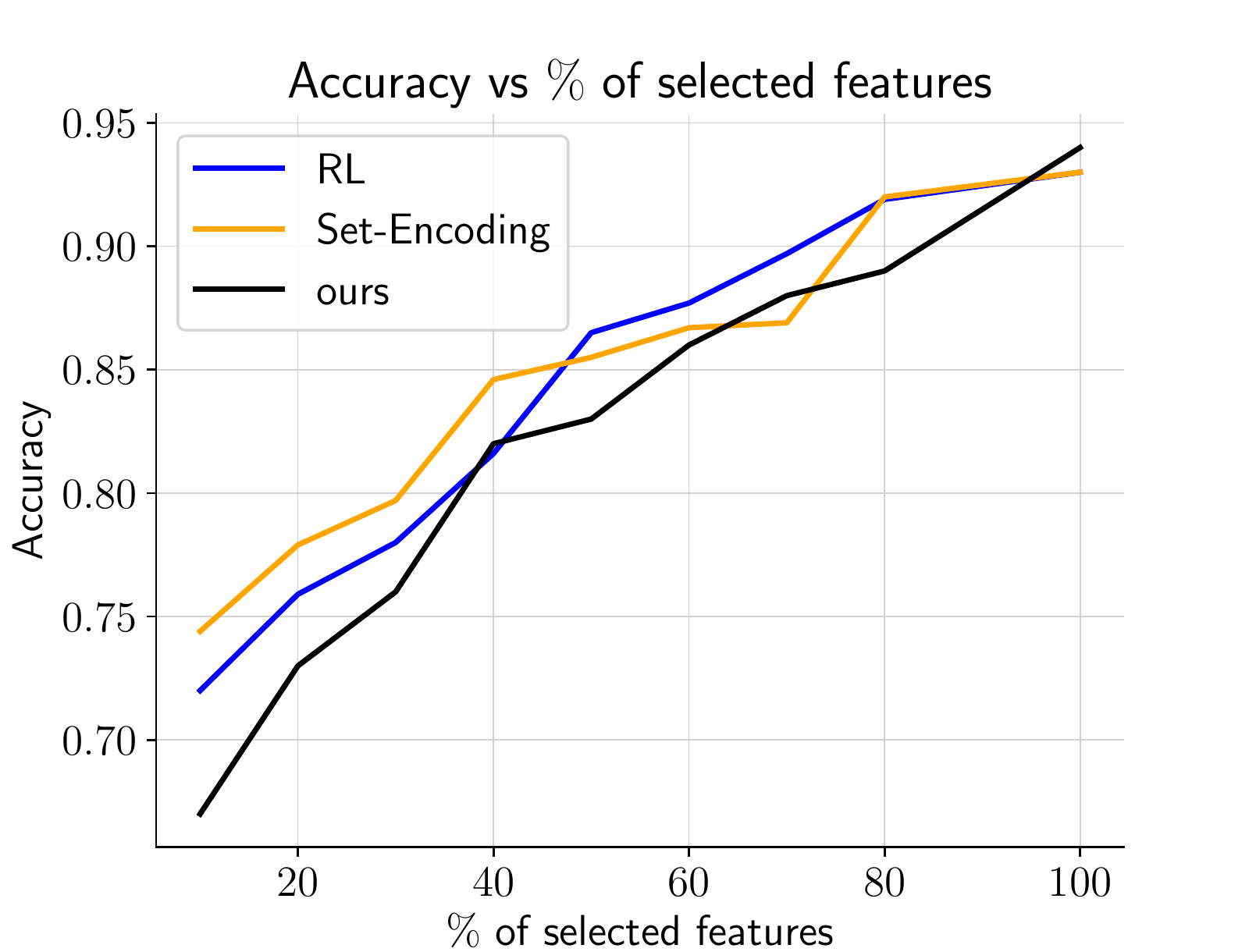}
    \caption{Forest dataset. The CWCF method is referred to as RL in this graph, while Set-encoding, another RL algorithm, is given a specific name.}
    \label{fig:forest}
\end{figure}
The dataset utilized in this section is known as the Forest dataset, which is  employed for the classification of pixels into seven distinct forest cover types. Each sample within the dataset corresponds to a 30 x 30 meter area of land. With a comprehensive collection of 54 features including elevation, aspect, slope, hillshade, soil type and more. The dataset encompasses a total of 581,012 samples.

Figure \ref{fig:forest} presents the classification accuracy, demonstrating that the proposed approach exhibits comparable performance to two other reinforcement learning (RL) based approaches. Notably, given the substantial size of the forest dataset, RL methods tend to yield superior performance; however, our approach remains competitive in terms of accuracy. It is noteworthy to mention that our approach requires zero training time, in contrast to the other approaches, which necessitate significant training durations. The subsequent section will delve into a detailed analysis of the training times associated with all examined methods.

\subsection{Computational cost comparison}
In addition to outperforming contemporary feature selection techniques, our active feature acquisition method has a faster computational time.
\begin{table}[t]
	\centering
	\caption{Comparative analysis of the time cost of RL-based methods and our methods on three datasets. We report the average testing duration for each sample~(second).}
        \vspace{-3mm}
	\begin{tabular}{c|ccc}
		\hline
		Method & CUBE & Forest & Our Synthetic data  \\
		\hline
		RL~(CWCF)~\cite{cwcf} & $243.3$ & $272.1$ & $237.2$\\
		Set-Encoding~\cite{shim2018joint}& $47.8$ & $76.07$ & $42.5$   \\
            \hline
		  \textbf{Ours}& $\mathbf{0.0056}$ & $\mathbf{0.04}$ & $\mathbf{0.0092}$   \\
		\hline
	\end{tabular}
	\vspace{-3mm}
	\label{tab:time}
\end{table}
Table~\ref{tab:time} demonstrates that our straightforward method has a substantial time advantage.

\section{Conclusion}
\label{Conclusion}
The present study introduces a new strategy for active feature acquisition in classification assignments. This method entails the sequential identification of the most informative feature subset to attain optimal prediction performance during testing, while simultaneously minimising cost. The approach we propose employs a novel lazy model that exhibits superior speed and efficiency in comparison to current techniques, yet yields comparable levels of accuracy.

\bibliography{ref}

\end{document}